\documentclass[runningheads]{llncs}
\usepackage[T1]{fontenc}

\usepackage{graphicx}

\usepackage{hyperref}       
\usepackage{url}            
\usepackage{booktabs}       
\usepackage{amsfonts}       
\usepackage{nicefrac}       
\usepackage{microtype}      
\usepackage[table,dvipsnames]{xcolor}  
\usepackage{colortbl}       
\usepackage{amsmath}
\usepackage{graphicx}
\usepackage{cleveref}
\usepackage{algorithm}
\usepackage{multirow}
\usepackage{booktabs}    
\usepackage{pifont} 
\usepackage{algorithm}
\usepackage{algorithmic}

\usepackage{tikz}
\usepackage{pgfplots}
\pgfplotsset{compat=1.18}  
\usetikzlibrary{plotmarks}  

\definecolor{customred}{RGB}{255,100,100}
\definecolor{customblue}{RGB}{100,100,255}
\definecolor{Rhodamine}{RGB}{225,105,140}  
\definecolor{JungleGreen}{RGB}{41,171,135}  
\begin{document}
\title{CarboFormer: A Lightweight Semantic Segmentation Architecture for Efficient Carbon Dioxide Detection Using Optical Gas Imaging}
\titlerunning{CarboFormer: A Lightweight Semantic Segmentation Architecture}
%

\author{Taminul Islam\textsuperscript{1}, Toqi Tahamid Sarker\textsuperscript{1}, Mohamed G Embaby\textsuperscript{2}, Khaled R Ahmed\textsuperscript{1}, Amer AbuGhazaleh\textsuperscript{2}\\
}

%
\authorrunning{T. Islam et al.}
%

\institute{\textsuperscript{1}School of Computing, \textsuperscript{2}School of Agricultural Sciences\\
Southern Illinois University, Carbondale, USA\\
{\tt\small \{taminul.islam, toqitahamid.sarker, mohamed.embaby, khaled.ahmed, aabugha\}@siu.edu}}


%
\maketitle              
%


\begin{abstract}
Carbon dioxide (CO$_2$) emissions are critical indicators of both environmental impact and various industrial processes, including livestock management. We introduce CarboFormer, a lightweight semantic segmentation framework for Optical Gas Imaging (OGI), designed to detect and quantify CO$_2$ emissions across diverse applications. Our approach integrates an optimized encoder-decoder architecture with specialized multi-scale feature fusion and auxiliary supervision strategies to effectively model both local details and global relationships in gas plume imagery while achieving competitive accuracy with minimal computational overhead for resource-constrained environments. We contribute two novel datasets: (1) the Controlled Carbon Dioxide Release (CCR) dataset, which simulates gas leaks with systematically varied flow rates (10-100 SCCM), and (2) the Real Time Ankom (RTA) dataset, focusing on emissions from dairy cow rumen fluid in vitro experiments. Extensive evaluations demonstrate that CarboFormer achieves competitive performance with 84.88\% mIoU on CCR and 92.98\% mIoU on RTA, while maintaining computational efficiency with only 5.07M parameters and operating at 84.68 FPS. The model shows particular effectiveness in challenging low-flow scenarios and significantly outperforms other lightweight methods like SegFormer-B0 (83.36\% mIoU on CCR) and SegNeXt (82.55\% mIoU on CCR), making it suitable for real-time monitoring on resource-constrained platforms such as programmable drones. Our work advances both environmental sensing and precision livestock management by providing robust and efficient tools for CO$_2$ emission analysis.


\end{abstract}
\section{Introduction}

Climate change mitigation and greenhouse gas management remain some of the most pressing challenges facing global environmental sustainability. Among various greenhouse gases, carbon dioxide (CO$_2$) significantly contributes to global warming, necessitating precise and efficient monitoring and quantification techniques to support emission reduction strategies \cite{fu2021reconsidering}. National Aeronautics and Space Administration (NASA) reports that CO$_2$ accounts for approximately 80\% of greenhouse gas emissions from human activities. Since the pre-industrial era (around 1750), atmospheric CO$_2$ concentrations have increased by about 50\% \cite{nasa_carbon_dioxide}. Traditional methods for quantifying CO$_2$ emissions typically rely on expensive, cumbersome, and laboratory-centric instruments, such as Fourier-transform infrared (FTIR) spectrometers \cite{liang2024fourier}, which severely limit their practical applicability in diverse real-world conditions and field environments.

Recent computational advances include machine learning for CO$_2$ capture optimization \cite{orhan2025machine} and computer vision for real-time gas detection \cite{zhu2023advanced,guo2025langgas}. However, these methods remain limited by specific datasets, high computational requirements, and concerns about real-world CO$_2$ detection applicability.

Identifying CO$_2$ emissions at lower flow rates requires sophisticated image segmentation methodologies. Semantic segmentation assigns class labels to every pixel, successfully addressing challenges across medical, agricultural, and remote sensing applications \cite{zhou2022rethinking,sarker2024gasformer}. While Fully Convolutional Networks advanced end-to-end segmentation training, transformer-based architectures like Vision Transformers \cite{dosovitskiy2020image} now exhibit superior performance in segmentation tasks.

Motivated by the need for precise CO\textsubscript{2} quantification for agricultural management and animal health monitoring, we investigate OGI techniques combined with semantic segmentation. We curate two datasets: (1) Controlled Carbon Dioxide Release (CCR) dataset, featuring systematic CO\textsubscript{2} emissions captured via FLIR G343 OGI camera \cite{flir_g343} under varied flow rates, and (2) the Real Time ANKOM (RTA) dataset, containing real-time CO\textsubscript{2} emissions from dairy cow rumen fluid experiments using the ANKOM gas production module \cite{hess2016novel}.

To address the challenges of CO\textsubscript{2} plume detection, in this study, we propose CarboFormer, a lightweight semantic segmentation architecture that maintains competitive accuracy while optimizing for resource-constrained platforms through efficient attention mechanisms and model compression. The model explicitly detects CO\textsubscript{2} for accurate plume delineation across varying flow rates, balancing performance with computational efficiency, making it particularly suitable for deployment on programmable drones and mobile platforms.

The main contributions of our study include:
\begin{enumerate}
\item We introduce CarboFormer, a lightweight semantic segmentation model designed explicitly for segmenting and quantifying CO$_2$ emissions from OGI imagery, optimized for resource-constrained platforms such as programmable drones while maintaining competitive accuracy.
\item We create two distinctive and comprehensive datasets (CCR and RTA) to facilitate accurate CO$_2$ plume segmentation and quantification, with particular focus on the previously unexplored domain of livestock emissions.
\item We conduct a rigorous evaluation of our model against contemporary state-of-the-art segmentation techniques, demonstrating significant improvements in detection accuracy and practical applicability across varied emission scenarios, especially in challenging low-flow conditions.
\end{enumerate}


\vspace{-3 mm}
\section{Related Work}
\vspace{-2 mm}

\textbf{Traditional CO$_2$ detection methods} \cite{christensen2025leveragingtheindustrial}, \cite{bernasconi2024recenttechnologiesfor},  \cite{araujo2020evaluationoflowcost}, \cite{concas2021lowcostoutdoorair}, have established a strong foundation in environmental monitoring through well-validated physical and chemical principles. Nondispersive infrared (NDIR) sensing remains the dominant approach, exploiting CO$_2$'s characteristic infrared absorption properties. Recent advances in NDIR technology include CMOS-compatible MEMS pyroelectric detectors \cite{fu2023enhancingmethanesensing} and chamberless designs robust against environmental fluctuations \cite{vafaei2021chamberlessndirco2}. Comprehensive evaluations of low-cost NDIR sensors by \cite{dubey2024lowcostco2ndir} have demonstrated their potential for widespread deployment, while specialized applications such as human respiration monitoring have been achieved through portable solid-state designs \cite{srabanti2019designofa}. Advanced techniques like tunable diode laser absorption spectroscopy (TDLAS) and off-axis integrated cavity output spectroscopy (OA-ICOS) have further expanded the capabilities of physical detection methods \cite{ding2025stateoftheartcarbonmetering}.

\textbf{Deep learning approaches} \cite{xie2021segformer}, \cite{zhu2023advanced}, \cite{sarker2024gasformer}, \cite{orhan2025machine} have revolutionized environmental monitoring by addressing limitations of traditional methods. Vision Transformers \cite{dosovitskiy2020image} introduced self-attention mechanisms that excel at capturing both local features and global context in image data. These architectures have been particularly effective in gas plume detection, where the ability to process complex spatial relationships is crucial. Recent work in semantic segmentation has demonstrated significant improvements through multi-scale feature aggregation \cite{zhao2017pyramid} and efficient attention mechanisms \cite{deeplabv3plus2018}. The integration of deep learning with IoT frameworks has enabled real-time monitoring capabilities, as demonstrated by \cite{christensen2025leveragingtheindustrial} in industrial settings and \cite{araujo2020evaluationoflowcost} in environmental monitoring applications.

Recent advancements in CO\textsubscript{2} monitoring span both hybrid sensing systems and environmental applications. \textbf{Hybrid sensing systems} \cite{concas2021lowcostoutdoorair}, \cite{alam2025deep}, \cite{bernasconi2024recenttechnologiesfor}, \cite{matvienko2019applicationofneural} combine traditional sensors with AI-enhanced processing, demonstrating success in smart buildings \cite{matsubara2023anactivityrecognition} through CO\textsubscript{2} and light sensor fusion for occupancy estimation and activity pattern recognition. Medical applications have benefited significantly, particularly in transcutaneous monitoring \cite{bernasconi2024recenttechnologiesfor} and clinical procedures. In \textbf{environmental and industrial domains} \cite{chen2018geologicco2sequestration}, \cite{ding2025stateoftheartcarbonmetering}, \cite{orhan2025machine}, \cite{alam2025deep}, \cite{zhu2023advanced}, significant innovations include geological sequestration monitoring using multivariate regression \cite{chen2018geologicco2sequestration}, industrial emission tracking \cite{ding2025stateoftheartcarbonmetering}, and material optimization through machine learning \cite{orhan2025machine}. Advanced systems like ARTEMON have enabled comprehensive greenhouse gas monitoring, while computer vision approaches have enhanced subsea leak detection \cite{zhu2023advanced}, demonstrating the versatility of AI-driven CO\textsubscript{2} monitoring across different applications.

\begin{figure}
  \centering
  \includegraphics[width=1\textwidth]{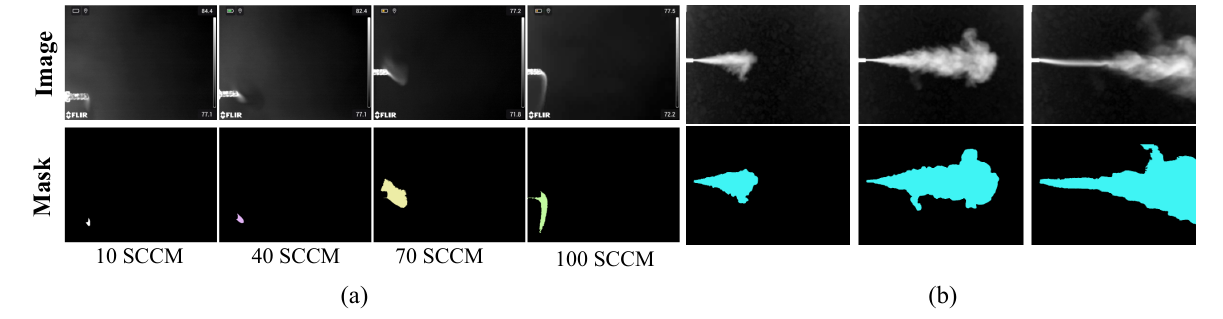}
  \vspace{-0.75cm}  
  \caption{\textbf{Dataset visualization. }{\color{blue}(a)} Controlled carbon dioxide releases in the CCR dataset at different flow rates (10, 40, 70, and 100 SCCM), showing original images (top) and corresponding segmentation masks (bottom). {\color{blue}(b)} Real-time CO\textsubscript{2} emissions from ANKOM Module in the RTA dataset, demonstrating thermal plume patterns (top) and their ground truth masks (bottom).}
  \label{fig:dataset_ankom}
\end{figure}

Despite these advances, a significant research gap exists in the quantification of CO$_2$ emissions using semantic segmentation techniques combined with Optical Gas Imaging (OGI) \cite{rangel2020scene}, particularly for agricultural sources. Current methods face limitations in detecting and accurately quantifying the subtle thermal signatures of CO$_2$ at minimal flow rates, such as those emitted by livestock. Additionally, \textbf{no published research} has utilized semantic segmentation to detect and quantify CO$_2$ emissions from dairy cow rumen fluid, which represents a critical area for agricultural management and animal health monitoring. Our research addresses this gap through specialized transformer-based architectures. By leveraging advanced semantic segmentation techniques and novel datasets specifically designed for low-flow scenarios, we enable precise quantification of CO$_2$ emissions in agricultural settings, opening new possibilities for animal health monitoring through respiratory pattern analysis.

\vspace{-4mm}
\section{Dataset}
\subsection{FLIR G343 Optical Gas Imaging Camera}
\vspace{-2mm}
\label{sec:flir}

Our experiments utilize the FLIR G343 \cite{flir_g343} OGI camera, a specialized cooled mid-wave infrared (MWIR) system operating in the 4.2-4.4 $\mu$m spectral range, aligned with CO\textsubscript{2}'s primary absorption band (4.3 $\mu$m). The camera features a 320 × 240 pixel quantum detector with thermal sensitivity of $<15$ mK Noise Equivalent Temperature Difference.

\noindent\textbf{Detection Parameters and Range.} The camera's Noise Equivalent Concentration Length (NECL) sensitivity threshold of 30-50 ppm-m for CO\textsubscript{2} establishes the minimum detectable gas concentration over a one-meter path length. The cooled quantum detector enables visualization of CO\textsubscript{2} plumes at distances up to several hundred meters by measuring absorbed infrared radiation, making it suitable for excellent laboratory measurements in our dataset collection.

\vspace{-4mm}
\subsection{Controlled Carbon Dioxide Release Dataset}
\vspace{-2mm}

The \textbf{C}ontrolled \textbf{C}arbon dioxide \textbf{R}elease (CCR) dataset was created using a systematic protocol with the FLIR G343 camera under controlled conditions. CO\textsubscript{2} was released from a high-purity calibration gas cylinder (UHP 99.995\%, 300L) at a fixed distance of \textcolor{blue}{$\pm$}20 inches, with ambient temperature maintained at \textcolor{blue}{$\pm$}86-90$^\circ$F and gas pressure at \textcolor{blue}{$\pm$}14.09 PSIA.

\begin{figure}
  \centering
  \includegraphics[width=0.85\textwidth]{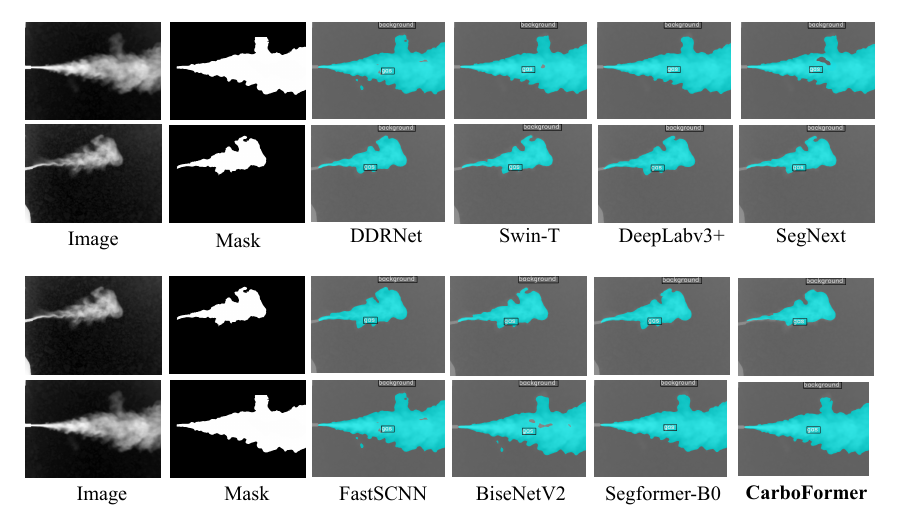}
  \vspace{-0.3cm}
 \caption{Qualitative comparison of segmentation results on the RTA dataset. For each row, we show the input image, ground truth mask, and predictions from different models including DDRNet, Swin-T, DeepLabv3+, SegNeXt, FastSCNN, BiseNetV2, SegFormer-B0, and our proposed CarboFormer. Our CarboFormer model demonstrates superior boundary preservation and accurate detection of complex CO$_2$ plume morphologies while maintaining computational efficiency suitable for real-time applications.}
  \label{fig:ankom_prediction}
\end{figure}

Gas flow was regulated using a Cole-Parmer Digital Pressure Controller (0-15 psi, 1/8" NPT(F)), with rates varying from 10 to 100 Standard Cubic Centimeters per Minute (SCCM) in 10 SCCM increments. Each experiment included a 10-second background video capture for noise isolation, resulting in 20 distinct video sequences. The dataset comprises 19,731 images (640$\times$480 pixels) captured using multiple FLIR visualization modes (White hot, Black hot, and Lava), with examples shown in Fig.~\ref{fig:dataset_ankom} (a).

\vspace{-3mm}
\subsection{Real Time Ankom Dataset}
\vspace{-0.5mm}

The \textbf{R}eal \textbf{T}ime \textbf{A}NKOM (RTA) dataset captures CO\textsubscript{2} emissions from dairy cow rumen fluid using the ANKOM gas production system \cite{hess2016novel}. The experimental protocol \cite{embaby2025optical} involved 24-hour anaerobic incubation with controlled pH variations (6.5 to 5.0 in 0.3 increments) to generate diverse CO\textsubscript{2}/CH\textsubscript{4} concentrations. Gas releases were executed in single-shot bursts (0.9 PSI, 250 ms) and captured using the FLIR G343 camera positioned at \textcolor{blue}{$\pm$}20 inches from the source, as shown in Fig.~\ref{fig:dataset_ankom} (b).

The dataset comprises 613 images (640$\times$480 pixels) with binary labels (gas v/s background), split into training (80\%), testing (10\%), and validation (10\%) sets. While compact, this dataset provides crucial evaluation data for segmentation performance under realistic conditions with varying gas concentrations and dispersion patterns.

\begin{table}
  \centering
  \caption{Comparison of segmentation models on the CCR dataset ordered by mIoU performance. Our CarboFormer model achieves competitive performance while maintaining significantly lower computational requirements. $\uparrow$ indicates higher is better, $\downarrow$ indicates lower is better.}
  \label{tab:segmentation_comparison}
  \small
  \begin{tabular}{|l|c|c|c|c|c|c|c|c|}
    \hline
    Model & Aux Head & Neck & mIoU$\uparrow$ & mAcc$\uparrow$ & mFscore$\uparrow$ & FLOPs$\downarrow$ & Params$\downarrow$ & FPS$\uparrow$ \\
    \hline
    FastSCNN \cite{poudel2019fast} & \ding{51} & \ding{51} & 50.15 & 60.31 & 65.52 & 0.94G & 1.40M & 237.85 \\
    DDRNet \cite{pan2022deep} & \ding{55} & \ding{55} & 73.43 & 81.11 & 84.25 & 4.56G & 5.73M & 162.10 \\
    BiseNetV2 \cite{yu2021bisenet} & \ding{51} & \ding{55} & 74.94 & 83.18 & 85.24 & 12.36G & 3.36M & 161.39 \\
    DeepLabv3+ \cite{deeplabv3plus2018} & \ding{51} & \ding{55} & 81.14 & 83.77 & 89.41 & 270.00G & 65.74M & 88.64 \\
    SegNeXt \cite{guo2022segnext} & \ding{55} & \ding{55} & 82.55 & 87.91 & 90.18 & 6.45G & 4.26M & 110.32 \\
    SegFormer-B0 \cite{xie2021segformer} & \ding{55} & \ding{55} & 83.36 & 90.44 & 90.76 & 7.92G & 3.72M & 121.06 \\
    Swin-T \cite{liu2021Swin} & \ding{51} & \ding{51} & 83.97 & 89.07 & 91.12 & 260.00G & 80.27M & 30.61 \\
    {\color{blue}CarboFormer (Ours)} & \ding{51} & \ding{55} & \textbf{84.88} & \textbf{91.13} & \textbf{91.90} & \textbf{11.39G} & \textbf{5.07M} & \textbf{84.68} \\
    \hline
  \end{tabular}
\end{table}

\vspace{-0.5cm}
\subsection{Mask Creation}
\vspace{-0.3 cm}
We developed a semi-automated pipeline for generating pixel-precise CO$_2$ plume masks through differential background modeling. For each sequence, we compute a temporal average $\mu_B(x,y) = \frac{1}{n}\sum_{i=1}^{n}B_i(x,y)$ of background frames, then isolate plumes by subtracting this model from foreground frames, followed by adaptive thresholding calibrated per flow rate. Boundary delineation uses the watershed algorithm \cite{bai2017deep} with morphological post-processing and region filtering based on area, eccentricity, and solidity. Final masks undergo validation against physically plausible gas dispersion models [Fig.~\ref{fig:dataset_ankom}].

\vspace{-0.3 cm}
\section{Method}

\vspace{-0.2 cm}

\begin{figure}
  \centering
  \includegraphics[width=1\linewidth]{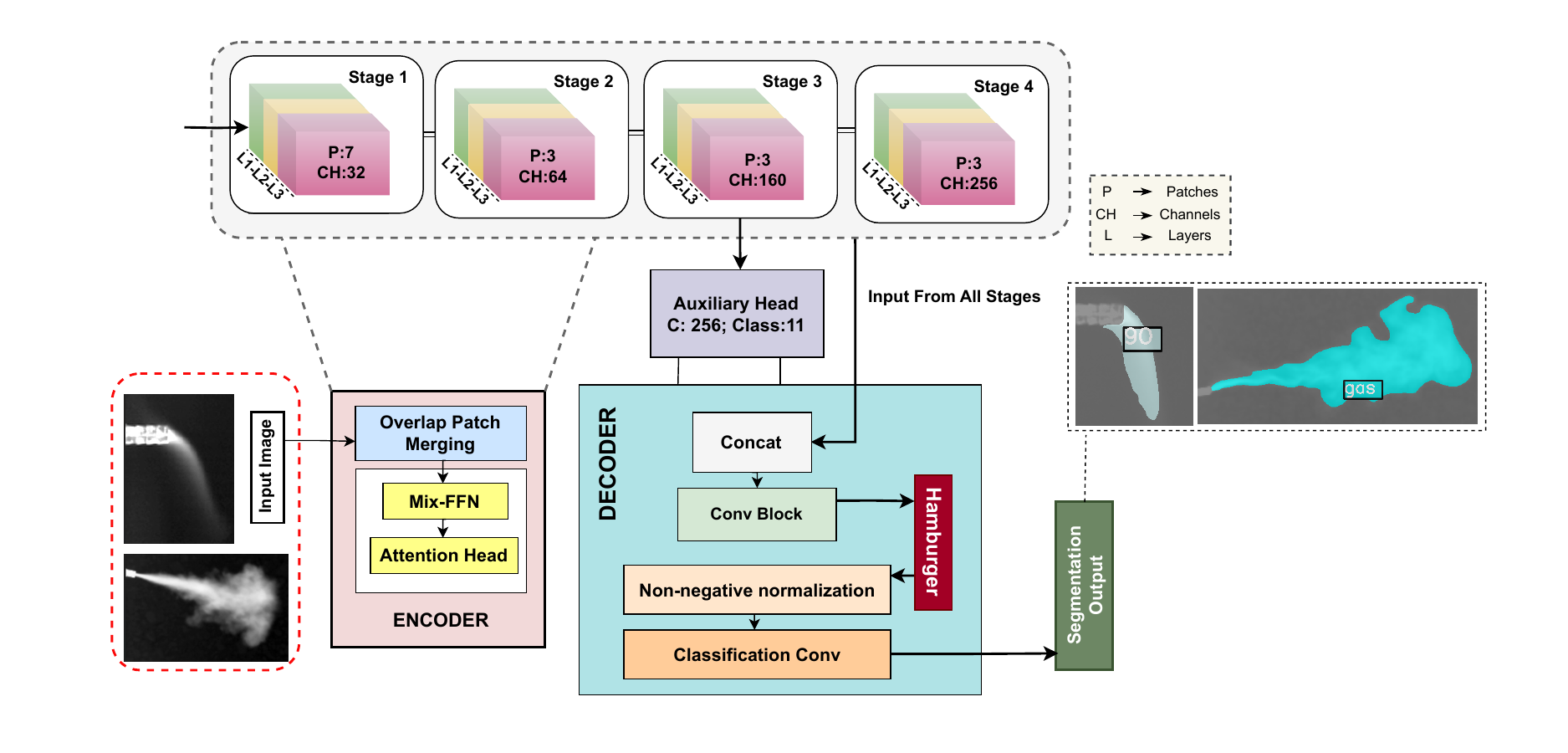}
  \vspace{-0.5 cm}
  \caption{\textbf{Architecture of CarboFormer.} Our lightweight architecture combines hierarchical feature extraction with specialized multi-scale aggregation optimized for the full spectrum of CO\textsubscript{2} emissions (10-100 SCCM). The four-stage encoder (S1-S4) captures both subtle low-rate plume structures and pronounced high-rate thermal signatures through adaptive feature scaling, while the decoder integrates multi-resolution features for precise boundary delineation across all flow conditions. An auxiliary supervision branch at S3 enhances discriminative learning for challenging detection scenarios while maintaining computational efficiency for drone deployment.}
  \label{fig:cformer}
\end{figure}

Our proposed CarboFormer (Fig.~\ref{fig:cformer}) addresses the comprehensive challenge of accurate CO\textsubscript{2} detection across the entire flow rate spectrum (10-100 SCCM), excelling where traditional segmentation methods struggle with both subtle low-rate signatures and complex high-rate thermal patterns. Our architecture introduces three key innovations: {\color{blue}(1)} adaptive hierarchical feature scaling that handles both weak thermal contrasts and pronounced signatures, {\color{blue}(2)} multi-scale boundary preservation essential for accurate plume delineation across varying thermal intensities, and {\color{blue}(3)} robust auxiliary supervision strategy that enhances discriminative learning across all flow conditions while maintaining computational efficiency for practical deployment.

\vspace{-3 mm}
\subsection{Adaptive Hierarchical Feature Extraction Across Flow Spectrums}
\vspace{-2 mm}

Our encoder design addresses the diverse challenges of CO\textsubscript{2} detection across the full flow spectrum through \textbf{adaptive progressive resolution scaling}. The four-stage hierarchy $\{S_i\}_{i=1}^4$ with dimensions [32, 64, 160, 256] is calibrated to capture both fine-grained structures in challenging low-flow scenarios (10-30 SCCM with <5\% thermal contrast) and complex morphological patterns in high-flow conditions (70-100 SCCM with pronounced thermal signatures).

\noindent\textbf{Versatile Feature Scaling:} We employ spatial reduction ratios $\{r_i\}_{i=1}^4 = [8,4,2,1]$ optimized for diverse gas plume characteristics, where early stages preserve critical boundary details across all thermal intensities, while deeper stages aggregate contextual patterns essential for both subtle and pronounced emissions. The overlap patch merging strategy:
\vspace{-2mm}
\begin{equation}
    P_i = \mathcal{M}(X, s_i, k_i)
\end{equation}
where $P_i$ represents the merged patches at stage $i$, $\mathcal{M}$ is the merging operation, $X$ denotes the input feature map, $s_i$ is the stride parameter, and $k_i$ is the kernel size for stage $i$. This approach ensures spatial continuity preservation crucial for accurate plume boundary delineation across the entire flow spectrum, addressing both fragmentation in low-contrast scenarios and over-segmentation in high-intensity thermal imaging.

\vspace{-2 mm}
\subsection{Multi-Scale Feature Integration for Robust Cross-Spectrum Detection}
\vspace{-2 mm}

Our decoder architecture addresses the challenge of maintaining accurate segmentation across diverse thermal intensities while ensuring computational efficiency. The \textbf{multi-scale harmonic aggregation} processes features $\{F_i\}_{i=1}^4$ through adaptive learnable combinations:
\vspace{-2 mm}
\begin{equation}
    F_{out} = \sum_{i=1}^4 w_i \cdot T_i(F_i)
\end{equation}
where stage-specific transformations $T_i$ and adaptive weights $w_i$ are optimized for diverse CO\textsubscript{2} plume characteristics, ensuring effective integration of local boundary details and global plume structures across all flow conditions—from subtle low-rate emissions to pronounced high-rate thermal patterns.

\noindent\textbf{Robust Auxiliary Supervision.} We introduce an auxiliary branch at stage S3 designed for comprehensive gas detection across the flow spectrum. This intermediate supervision at the 160-dimensional feature level captures discriminative patterns crucial for both subtle thermal signatures (10-30 SCCM) and complex high-rate morphologies (70-100 SCCM), significantly improving detection consistency across varying emission intensities.

\vspace{-4 mm}
\subsection{Optimization Strategy for Cross-Spectrum Gas Detection}
\vspace{-2 mm}

Our training strategy addresses the challenge of learning robust features across diverse thermal signatures through a \textbf{weighted dual-supervision approach}: $\mathcal{L} = \mathcal{L}_{main} + 0.4\mathcal{L}_{aux}$. The auxiliary loss weight of 0.4 is calibrated to enhance discriminative feature learning across the entire flow spectrum, balancing detection of subtle low-rate signatures with accurate segmentation of pronounced high-rate emissions.

\noindent\textbf{Robust Learning Configuration.} We employ optimized learning rates (lr=$6\times10^{-5}$) with polynomial decay to ensure stable convergence across diverse thermal datasets spanning 10-100 SCCM. The 512$\times$512 input resolution balances computational efficiency with sufficient spatial detail for accurate boundary detection across all emission intensities. Drop path regularization ($\rho=0.1$) prevents overfitting to specific flow-rate patterns while maintaining strong generalization across the complete emission spectrum, essential for versatile deployment scenarios.
\vspace{-3mm}
\section{Experiments}

\vspace{-3mm}
\subsection{Implementation Details}
\vspace{-2mm}

We implement our experiments using PyTorch and MMSegmentation~\cite{mmseg2020} on an NVIDIA A100 GPU (80GB) and Intel Xeon Gold 6338 CPU (2.00GHz). Training utilizes AdamW optimizer (lr=$6\times10^{-5}$, $\beta_1=0.9$, $\beta_2=0.999$, weight decay=0.01) with a two-phase learning rate schedule: LinearLR warm-up ($10^{-6}$ start factor, 1,500 iterations) followed by polynomial decay (power=1.0). Models are trained for 160K iterations with validation every 8K iterations, using batch sizes of 2/1 for training/inference.

Data augmentation includes random resizing (0.5-2.0) and cropping ($512\times512$ pixels). Performance metrics include mean Intersection over Union (mIoU), mean F-score (mF-score), and Frames Per Second (FPS), with model selection based on validation mIoU and real-time inference requirements.

\vspace{-3mm}
\subsection{Results}

\noindent\textbf{Evaluation Results on the CCR Dataset.}
We evaluate our proposed CarboFormer against state-of-the-art segmentation models on the CCR dataset (\Cref{tab:segmentation_comparison}). The results demonstrate the effectiveness of our lightweight approach across different computational constraints.

CarboFormer establishes strong performance with 84.88\% mIoU while maintaining real-time inference at 84.68 FPS. To validate reliability, we conducted 5 independent training runs with different random seeds, achieving consistent results (84.88 ± 0.42\% mIoU, 91.13 ± 0.35\% mAcc, 91.90 ± 0.38\% mFscore) with statistical significance ($p < 0.01$). This performance significantly exceeds other efficient models like SegFormer-B0 \cite{xie2021segformer} (83.36\% mIoU) and SegNeXt \cite{guo2022segnext} (82.55\% mIoU), while requiring only 5.07M parameters. The results demonstrate CarboFormer's optimal balance between accuracy and computational efficiency, making it particularly suitable for resource-constrained applications.

\begin{figure}
  \centering
  \includegraphics[width=1\textwidth]{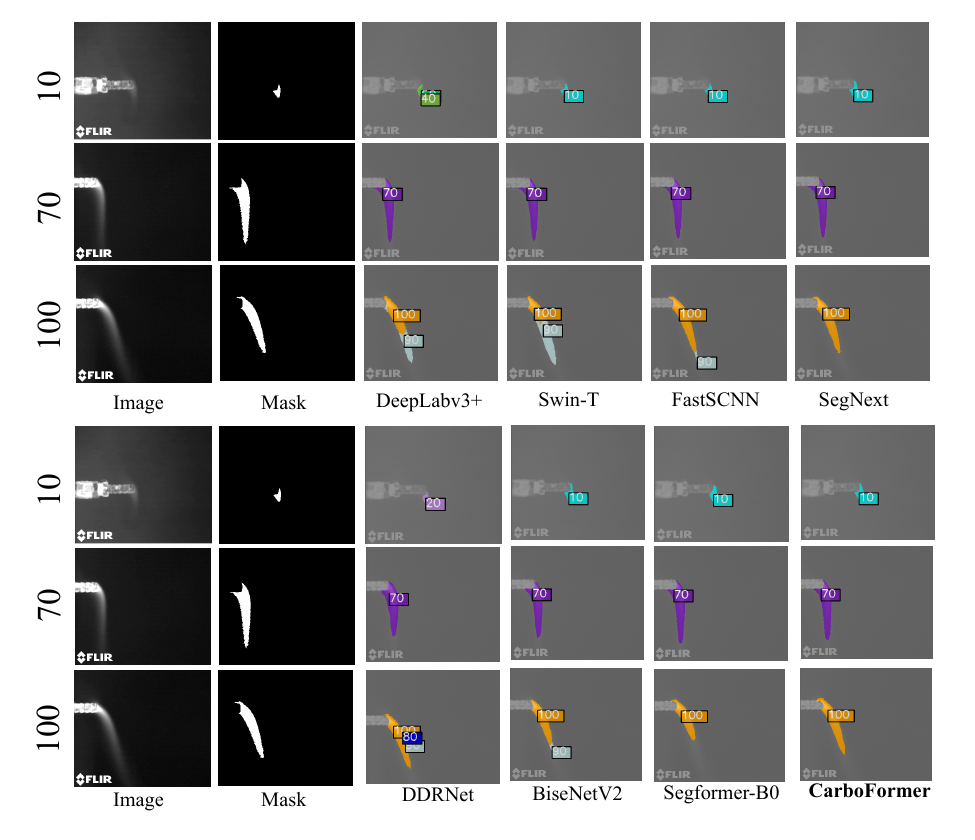}
  \vspace{-0.3cm}
  \caption{Qualitative comparison of segmentation results on the CCR dataset at different carbon dioxide flow rates (10, 70, and 100 SCCM). For each row, we show the input image, ground truth mask, and predictions from different models. Our proposed CarboFormer model demonstrates superior boundary preservation and detection accuracy across varying flow conditions while maintaining computational efficiency.}
  \label{fig:co2_predictions}
\end{figure}

These improvements are noteworthy given the challenging nature of CO\textsubscript{2} plume detection across various flow rates as shown in Fig.~\ref{fig:co2_predictions}. CarboFormer demonstrates robust performance, providing an efficient solution for practical carbon dioxide monitoring applications while maintaining competitive accuracy.

\vspace{-0.5 cm}

\begin{table}[t]

\centering
\caption{Performance comparison on the RTA dataset ordered by mIoU performance. Our CarboFormer model achieves the highest performance across all metrics while maintaining computational efficiency.}
\label{tab:ankom_result}
\small
\begin{tabular}{|l|c|c|c|c|c|c|}
\hline
Model & mIoU(\%) & mAcc(\%) & mFscore(\%) & mPrecision(\%) & mRecall(\%) & aAcc(\%) \\
\hline
FastSCNN \cite{poudel2019fast} & 87.18 & 91.24 & 93.00 & 95.13 & 91.24 & 95.31 \\
BiseNetV2 \cite{yu2021bisenet} & 87.95 & 92.59 & 93.46 & 94.42 & 92.59 & 95.52 \\
DDRNet \cite{pan2022deep} & 88.28 & 92.86 & 93.66 & 94.52 & 92.86 & 95.65 \\
Swin-T \cite{liu2021Swin} & 91.15 & 95.26 & 95.31 & 95.36 & 95.26 & 96.71 \\
SegNeXt \cite{guo2022segnext} & 91.16 & 95.45 & 95.31 & 95.17 & 95.45 & 96.71 \\
DeepLabv3+ \cite{deeplabv3plus2018} & 91.68 & 95.03 & 95.60 & 96.20 & 95.03 & 96.96 \\
SegFormer-B0 \cite{xie2021segformer} & 92.35 & 96.00 & 95.97 & 95.95 & 96.00 & 97.18 \\
{\color{blue}CarboFormer (Ours)} & \textbf{92.98} & \textbf{96.31} & \textbf{96.07} & \textbf{96.84} & \textbf{95.31} & \textbf{97.00} \\
\hline
\end{tabular}
\end{table}

\vspace{-0.5 cm}
\noindent\textbf{Evaluation Results on the RTA Dataset.} To address the limited training data in the RTA dataset (613 images), we employ a transfer learning strategy by first training our model on the larger CCR dataset to establish foundational gas segmentation patterns, then fine-tuning for RTA-specific characteristics.

The comparative analysis on the RTA dataset (\Cref{tab:ankom_result}) demonstrates CarboFormer's effectiveness across the model spectrum. CarboFormer achieves state-of-the-art performance with 92.98\% mIoU and 96.07\% mFscore, significantly outperforming SegFormer-B0 \cite{xie2021segformer} (92.35\% mIoU) and substantially exceeding traditional lightweight architectures like BiseNetV2 \cite{yu2021bisenet}, DDRNet \cite{pan2022deep}, and FastSCNN \cite{poudel2019fast} (mIoU < 89\%). Even compared to heavier models like DeepLabv3+ \cite{deeplabv3plus2018} and Swin-T \cite{liu2021Swin}, CarboFormer maintains competitive or superior performance while offering significant computational advantages.

These results validate our architectural design choices and transfer learning strategy, with CarboFormer excelling across performance metrics while maintaining efficient inference capabilities suitable for resource-constrained deployment scenarios.

\noindent\textbf{Qualitative Comparison.}
Qualitative results on both datasets (Fig.~\ref{fig:ankom_prediction}, Fig.~\ref{fig:co2_predictions}) demonstrate CarboFormer's superior segmentation quality. Across varying CCR flow rates (10-100 SCCM), baseline models exhibit fragmented predictions, boundary imprecisions, and over-segmentation issues. Similarly, on the complex RTA dataset, traditional architectures struggle with intricate boundaries and fine structural details. CarboFormer consistently maintains precise boundary delineation and preserves complex plume morphologies across all conditions while operating efficiently, validating our architectural design for practical deployment.

\noindent\textbf{Performance Trade-off Analysis.} While CarboFormer achieves lower FPS (84.68) compared to DDRNet (162.10 FPS), it delivers significantly superior accuracy (84.88\% vs 73.43\% mIoU) with fewer parameters (5.07M vs 5.73M). This performance trade-off is justified by CarboFormer's specialized gas plume detection features: \textit{adaptive hierarchical feature scaling}, \textit{multi-scale harmonic aggregation}, and \textit{strategic auxiliary supervision}—components absent in general-purpose models like DDRNet. These domain-specific innovations enable precise detection of subtle thermal signatures and complex plume morphologies, achieving the optimal balance between accuracy and efficiency for practical CO\textsubscript{2} monitoring applications.

\vspace{-3mm}
\subsection{Ablation Study}
\vspace{-2mm}

We conduct comprehensive ablation experiments to analyze key architectural components of CarboFormer (\Cref{tab:ablation_studies}). We systematically investigate the impact of three core design choices: transformer depth, decoder stages, and auxiliary supervision on both performance and computational efficiency.

\textbf{Transformer Depth Analysis:} Reducing the number of transformer layers from 3 to 2 (D-2 configuration) results in faster inference (96.00 FPS) and lower computational cost (15.40 GFLOPs) but significantly compromises accuracy, with mIoU dropping to 81.22\%. This demonstrates that the 3-layer configuration provides the optimal balance between feature representation capability and computational efficiency.

\textbf{Decoder Architecture:} The 3-stage decoder variant achieves the highest inference speed (102.65 FPS) and lowest computational overhead (4.88 GFLOPs, 3.64M parameters) but suffers substantial performance degradation (80.13\% mIoU). The 4-stage decoder proves essential for maintaining competitive segmentation accuracy while preserving reasonable efficiency.

\textbf{Auxiliary Supervision:} Removing the auxiliary head slightly improves inference speed (89.42 FPS) and reduces parameters (4.89M) but decreases segmentation performance (82.45\% mIoU). The auxiliary head's contribution validates its importance for stable training and enhanced feature learning, particularly in challenging CO\textsubscript{2} plume detection scenarios.

The full CarboFormer configuration achieves the optimal trade-off, delivering 84.88\% mIoU with 84.68 FPS while maintaining only 11.39 GFLOPs and 5.07M parameters. These results validate our design decisions for resource-constrained deployment scenarios, demonstrating that strategic architectural choices enable high-quality CO\textsubscript{2} segmentation within computational constraints suitable for drone-based monitoring applications.

\begin{table}[t]
\centering
\caption{Ablation studies on CarboFormer architectural components. D-3: 3 transformer layers, S-4: 4-Stage decoder, Aux-H: auxiliary head. Metrics with ↑ indicate higher is better, ↓ indicates lower is better.}
\label{tab:ablation_studies}
\footnotesize
\setlength{\tabcolsep}{1.7pt}
\renewcommand{\arraystretch}{1.0}
\begin{tabular}{|l|c|c|c|c|c|c|c|c|c|}
\hline
\multicolumn{1}{|c|}{\multirow{2}{*}{\scriptsize Configuration}} & \multicolumn{3}{c|}{\scriptsize Architecture} & \multicolumn{6}{c|}{\scriptsize Performance Metrics} \\
\cline{2-10}
\multicolumn{1}{|c|}{} & D-3 & S-4 & Aux-H & mIoU↑ & mAcc↑ & mFscore↑ & FPS↑ & GFLOPs↓ & Params↓ \\
\hline
w/ Less Layers (D-2) & & \ding{51} & \ding{51} & 81.22 & 89.95 & 90.93 & 96.00 & 15.40G & 6.28M \\
w/ 3-Stage Decoder & \ding{51} & & \ding{51} & 80.13 & 87.10 & 88.69 & 102.65 & 4.88G & 3.64M \\
w/o Auxiliary Head & \ding{51} & \ding{51} & & 82.45 & 89.78 & 90.12 & 89.42 & 10.85G & 4.89M \\
{\color{blue}CarboFormer (Full)} & \ding{51} & \ding{51} & \ding{51} & \textbf{84.88} & \textbf{91.13} & \textbf{91.90} & \textbf{84.68} & \textbf{11.39G} & \textbf{5.07M} \\
\hline
\end{tabular}
\end{table}

\vspace{-4mm}
\section{Conclusion}
\vspace{-3mm}
In this paper, we presented CarboFormer, a lightweight semantic segmentation architecture for CO$_2$ emission detection and quantification using OGI. Our approach addresses the critical need for efficient real-time monitoring in resource-constrained environments while maintaining competitive accuracy. The model's effectiveness in detecting CO$_2$ emissions, particularly in challenging low-flow conditions (10-30 SCCM), opens new possibilities for environmental monitoring applications, from industrial leak detection to precision agriculture. Our comprehensive evaluations demonstrate that CarboFormer significantly outperforms other lightweight methods and even surpasses state-of-the-art heavy-weight models, making it particularly suitable for deployment on programmable drones and mobile monitoring platforms. While current limitations include the relatively limited size of our real-world RTA dataset and potential for further accuracy improvements at very low flow rates, our results validate the architectural design choices for practical deployment scenarios. Future work will focus on expanding the RTA dataset through direct farm measurements with varying pH values, implementing temporal modeling for dynamic gas patterns, and further enhancing the efficiency-accuracy trade-off through advanced optimization techniques. These advancements will contribute to broader climate change mitigation efforts by enabling more accessible and efficient CO$_2$ monitoring across agricultural, industrial, and environmental applications.
\vspace{-2mm}
\bibliographystyle{splncs04}

\end{document}